\documentclass[lettersize,journal]{IEEEtran}
\usepackage{amsmath,amsfonts}
\usepackage{algorithmic}
\usepackage{algorithm}
\usepackage{array}
\usepackage[caption=false,font=normalsize,labelfont=sf,textfont=sf]{subfig}
\usepackage{textcomp}
\IfFileExists{stfloats.sty}{\usepackage{stfloats}}{}
\usepackage{url}
\usepackage{verbatim}
\usepackage{graphicx}
\usepackage{cite}
\begin{document}

\title{Accelerating Robot Path Planning via Connectivity-Preserving Region Proposal Network}

\author{Zhanzheng Ma, Cancan Zhao, Shuai Zhang, and Bo Ouyang%
\thanks{This work was supported by the National Key Research and Development Program of China (Grant No. 2023YFB4706000), in part by the National Natural Science Foundation of China (Grant No. 52305018), in part by the Basic Science Centre Program of the National Natural Science Foundation of China (Grant No. 72188101).}%
\thanks{Zhanzheng Ma, Cancan Zhao, Shuai Zhang, and Bo Ouyang are with the School of Management, Hefei University of Technology, Hefei 230009, China (e-mail: 2023171324@mail.hfut.edu.cn; 2021110733@mail.hfut.edu.cn; 2021010078@mail.hfut.edu.cn; boouyang@hfut.edu.cn).}%
\thanks{Corresponding author: Bo Ouyang.}%
}

\maketitle

\begin{abstract}
Mobile robot path planning methods are often constrained by vast search spaces, resulting in latency in sampling-based algorithms. Learning-based approaches frequently suffer from local region fragmentation and global topological inconsistency. To tackle the problem, we present the Connectivity-Preserving Region Proposal Network (CP-RPN), a segmentation-guided model designed to predict compact and topologically connected candidate regions, significantly compressing the search space. Specifically, we design a segmentation model that leverages a Deformable Attention Transformer (DAT) to capture long-range dependencies for global connectivity, with a Deconvolutional decoder to preserve fine-grained spatial details. To guarantee the connectivity of the predicted mask, we design a composite loss function that combines Cross-Entropy loss for pixel-wise supervision, a Connectivity-Aware loss to enhance local coherence, and a Topological Continuity loss based on persistent homology to enforce global connectivity. Building on these high-connectivity corridor-like regions, the Voronoi diagram is used to plan the path, backed by a local A* fallback mechanism to ensure robustness. Experimental results demonstrate that CP-RPN reduces the candidate region size by over 60.13\% compared to the MPT baseline and achieves deterministic low-latency planning (avg. 0.11s) with a 99.60\% success rate, outperforming traditional sampling-based algorithms in stability.
\end{abstract}


\section{Introduction}
\label{sec:introduction}
\IEEEPARstart{P}{ath} planning constitutes a fundamental pillar of mobile robot autonomy, particularly in safety-critical navigation scenarios. The primary objective is to compute a collision-free trajectory that connects a starting position to a target point, often within large-scale environments characterized by dense obstacles and complex topologies \cite{lavalle2006planning}, \cite{elbanhawi2014sampling}. Classical solutions are predominantly categorized into deterministic graph-search algorithms (e.g., A* \cite{hart1968formal}, Dijkstra \cite{dijkstra1959note}), probabilistic sampling-based strategies (e.g., RRT \cite{LaValle1998RapidlyexploringRT}, PRM \cite{kavraki1996probabilistic}), and local trajectory optimization methods (e.g., CHOMP \cite{Ratliff2009CHOMPGO}, MPC \cite{garcia1989model}). However, these approaches are constrained by inherent scalability issues. Graph-based methods suffer from computational intractability in large-scale environments, whereas sampling-based planners are prone to stochastic instability and varying convergence rates, especially within narrow passages in cluttered spaces \cite{elbanhawi2014sampling}.

\begin{figure}[!t]
    \centering
    \includegraphics[width=\linewidth,trim={0cm 0cm 0cm 0cm},clip]{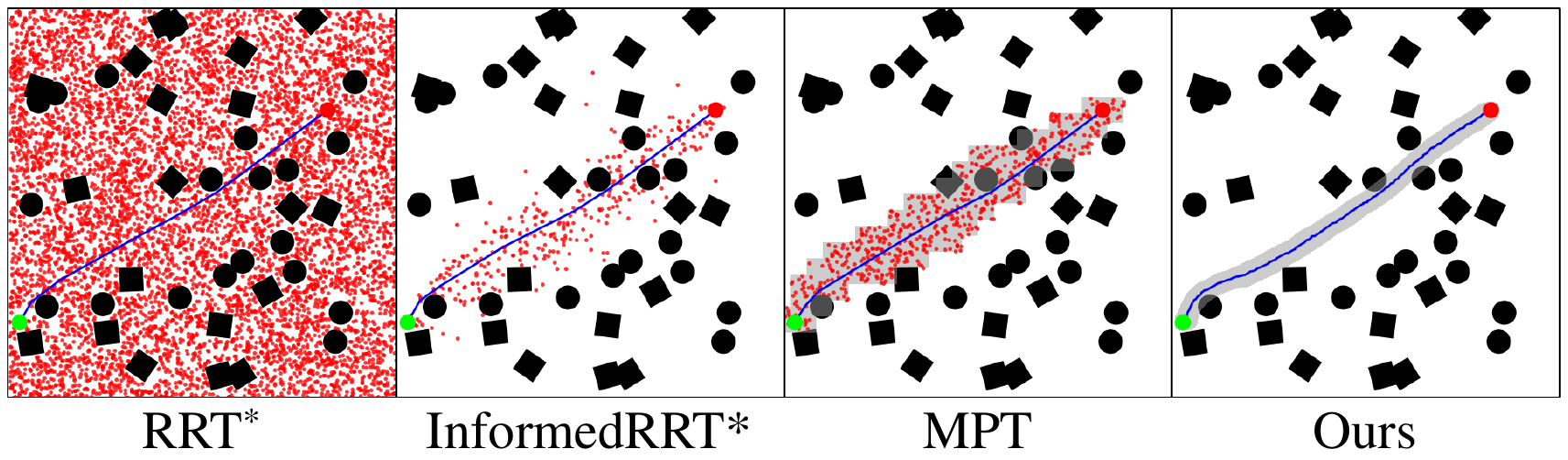}
    \vspace{-0.6cm}
    \caption{The search process of RRT*, Informed RRT*, MPT, and our method. The red points represent the explored vertices, with the large red point indicating the start, the green point marking the goal, and the blue line representing the final path. Ours extracts a highly compact candidate region. This allows the Voronoi planner to avoid redundant sampling.}
    \label{fig_region_proposal}
    \vspace{-0.4cm}
\end{figure}

To overcome these challenges, learning-based motion planning has emerged as a promising paradigm. By utilizing deep architectures such as Convolutional Neural Networks (CNNs) and Multi-Layer Perceptrons (MLPs) to encode environmental priors, these approaches either learn heuristic functions to guide sampling \cite{Wang2020NeuralRL,yonetani2021pathplanningusingneural} or directly generate trajectory sequences \cite{Qureshi2018MotionPN}. This enables the planner to focus on high-probability states rather than exploring the entire space, significantly accelerating convergence \cite{sanchez2021survey}. However, these methods frequently struggle to capture long-range spatial dependencies in complex environments, often lacking a comprehensive global topological perspective due to the limited receptive fields inherent in convolutional architectures. Consequently, such topological defects render the learned guidance spatially incoherent, often trapping the planner in local minima or false dead-ends, which severely undermines the intended efficiency. To mitigate the limited receptive fields of CNNs, the Motion Planning Transformer (MPT) \cite{johnson2021motion} proposes a transformer-based \cite{Vaswani2017AttentionIA} region proposal network designed to capture long-term spatial dependencies among local map patches. By exploiting global self-attention, MPT explicitly models long-range spatial dependencies to restrict the search space for promising regions, thereby significantly reducing sampling complexity. However, MPT relies on fixed-size patches, which inherently limit its spatial resolution. This approach generates overly broad candidate regions to maintain topological connectivity, as illustrated in Fig.~\ref{fig_region_proposal}. These coarse masks inevitably cover a large number of obstacles. Consequently, the downstream planner must perform redundant collision checks within these inflated areas, significantly increasing the computational overhead.

\begin{figure*}[!t]
\centerline{\includegraphics[width=18cm,trim={0cm 2.1cm 0cm 0cm},clip]{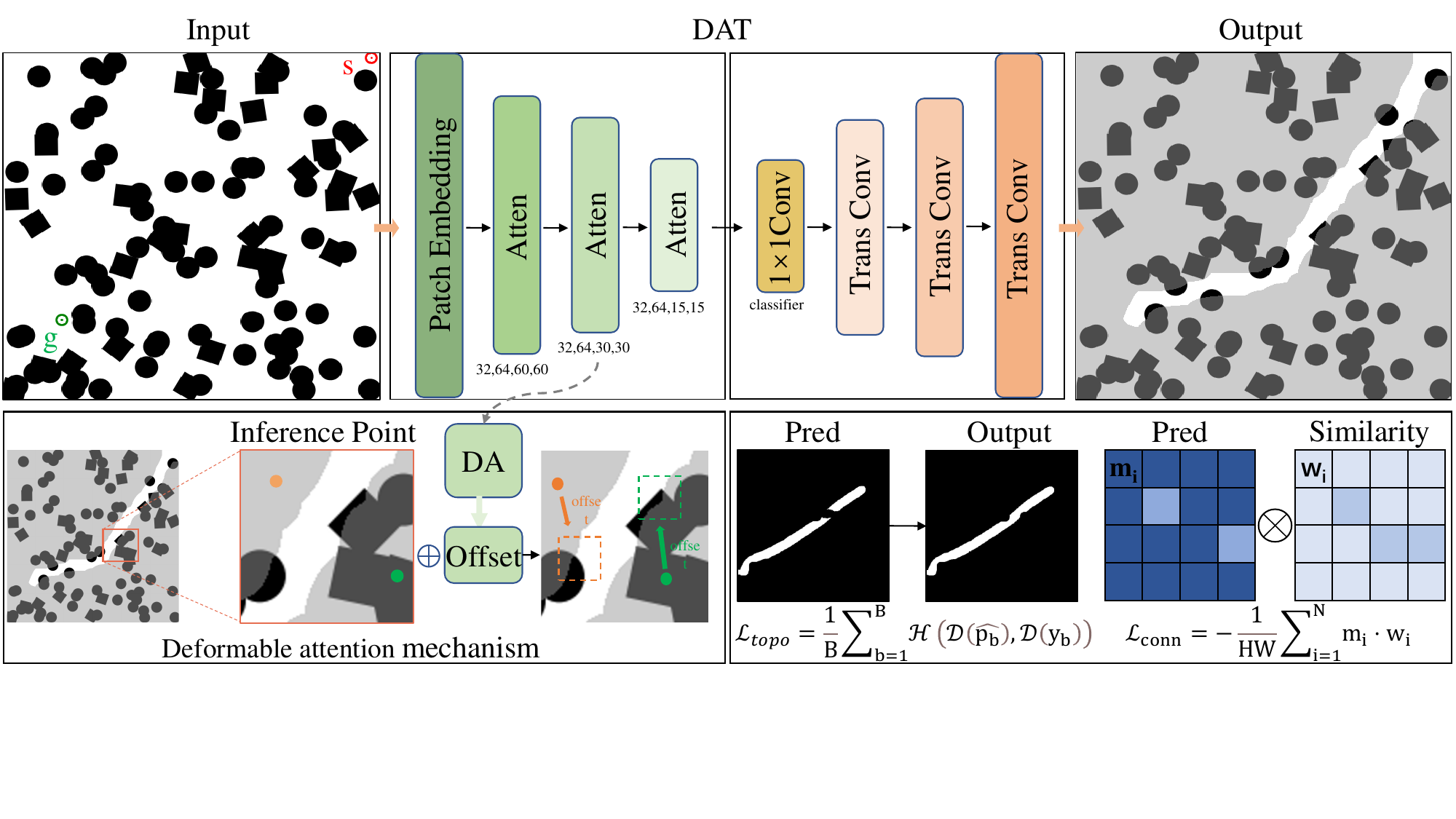}}
\vspace{-0.7cm}
\caption{The overall architecture of the proposed path planning network. The inputs are a binary grid map and a start/goal map. A DAT encoder and a deconvolutional decoder are used to predict a path region. A novel Connectivity-Aware Loss is applied to ensure the connection of the output.}
\label{fig_pipelne}
\vspace{-0.4cm}
\end{figure*}

To address these issues, this paper transforms the candidate region prediction problem into an image segmentation task to capture both local and global information, thereby effectively reducing the candidate region size. Specifically, we design an encoder-decoder architecture inspired by the U-Net \cite{ronneberger2015unet} that combines the DAT \cite{xia2022vision} with the deconvolutional decoder \cite{7410535} to predict connected regions that contain the optimal path. The DAT serves as a powerful encoder that captures long-range dependencies and global topological connectivity, thereby ensuring the structural integrity of the predicted path region. Meanwhile, a deconvolutional decoder utilizes stacked transposed convolutions to preserve fine-grained spatial details during upsampling, which are crucial for delineating precise path boundaries. It significantly improves the efficiency and accuracy of path planning by predicting the candidate region.  Given the high fidelity and topological connectivity of the candidate regions, we employ a Voronoi graph \cite{Choset1996SensorBM} in the narrow corridor-like region to determine the path, thereby reducing computing time. Additionally, to address connectivity issues arising from small-object segmentation in semantic segmentation tasks, we propose a composite loss that combines a connectivity-aware loss with a topological continuity loss. The connectivity-aware loss enhances the connectivity of the predicted segmentation map via local mean filtering, thereby making the path region more coherent. The topological continuity loss, using persistent homology techniques \cite{Edelsbrunner2009ComputationalTA} \cite{Qi2023DynamicSC}, ensures topological consistency between the predicted segmentation map and the ground truth labels \cite{Hu2019TopologyPreservingDI}, further improving the accuracy and stability of path planning.

In summary, the contributions of this study are as follows: 
\begin{enumerate}

\item 
We propose a novel method for candidate region prediction that integrates the DAT framework with a deconvolutional decoder, significantly reducing the search space for path planning algorithms.

\item 
We introduce a composite loss function comprising connectivity-aware, topological continuity, and cross-entropy terms. This formulation ensures effective path planning within connected and topologically consistent regions while preserving spatial compactness.

\item 
Experimental results demonstrate that our model achieves a success rate comparable to state-of-the-art methods while reducing the candidate path region size by 60.13\% and accelerating computation by 42.10\% compared to the baseline.
\end{enumerate}

\section{Related Works}

\subsection{Traditional Motion Planning}

Path planning is a fundamental task of mobile robot autonomy, dedicated to computing safe trajectories between start and goal configurations in complex workspaces. Classical approaches are broadly divided into graph-search and sampling-based algorithms. Graph-search methods such as the Dijkstra algorithm \cite{dijkstra1959note} and A* \cite{hart1968formal} guarantee optimality on discretised grids but suffer from exponential growth in computational cost as map resolution or obstacle complexity increases.
Sampling-based planners, including the Probabilistic Roadmap (PRM) \cite{kavraki1996probabilistic} and the Rapidly-exploring Random Tree (RRT) \cite{LaValle1998RapidlyexploringRT}, provide probabilistic completeness and scale better to high-dimensional spaces. Refinements such as RRT* \cite{Karaman2011SamplingbasedAF} and Informed RRT* \cite{gammell2014informed} improve asymptotic optimality through cost-aware sampling. However, these methods fundamentally rely on stochastic sampling strategies that explore the entire configuration space without structural guidance. This blind exploration results in excessive collision checking and stochastic latency, particularly in narrow passages or dense environments \cite{elbanhawi2014sampling}. Consequently, recent studies have turned to data-driven methods that employ environmental priors to compress the candidate region and speed up convergence.

\subsection{Learning-Based Path Planning}
Deep learning has reshaped the motion planning landscape by mitigating the computational bottlenecks of classical algorithms through policy learning, heuristic prediction, and end-to-end path generation. Initial methods focused on training neural networks to steer traditional planners toward promising areas by learning suitable sampling distributions or heuristics. Neural RRT* \cite{Wang2020NeuralRL} uses a CNN to predict the probability distribution of the optimal path. Ichter \textit{et al.}~\cite{Ichter2017LearningSD} employ a CVAE to learn a latent sampling distribution conditioned on the planning problem. Neural A* \cite{yonetani2021pathplanningusingneural} reformulates the search process as a differentiable module to refine node expansion priorities. Although these strategies enhance exploration efficiency, they still depend on incremental sampling and retain the scalability limits of classical frameworks. A more radical direction aims for end-to-end path prediction. The Motion Planning Network (MPNet) \cite{Qureshi2018MotionPN} exemplifies this trend, employing an encoder–decoder architecture that maps environmental representations to feasible trajectory sequences, achieving rapid inference in familiar domains. However, these direct-generation methods offer no theoretical guarantees of being collision-free and tend to generalise poorly to unseen situations, frequently breaking down in cluttered environments.

Region proposal methods emerge to resolve the conflict between computational speed and feasibility guarantees. Predicting exact trajectories is unreliable, while searching the entire space is inefficient. Therefore, recent paradigms predict compact navigable regions to constrain the downstream search space. Early CNN-based semantic segmentation models (e.g., \cite{Shelhamer2014FullyCN}) identify free space regions but lack global connectivity awareness due to local receptive fields. Transformer-based models overcome this limitation through long-range dependency modeling. For example, Motion Planning Transformer (MPT) \cite{johnson2021motion} predicts interconnected regions to efficiently guide RRT*. However, MPT relies on a patch-wise classification paradigm, which imposes an inherent limit on spatial resolution. By dividing the map into fixed-size blocks, MPT produces coarse and blocky regions that loosely approximate the true candidate space. These inflated regions inevitably encompass obstacles. The downstream planner is forced to perform redundant collision checks within these ambiguous areas. In contrast, CP-RPN formulates the region proposal as a pixel-wise semantic segmentation task. It extracts highly compact and connected candidate regions. This mechanism allows the downstream Voronoi planner to eliminate redundant sampling.

\subsection{Segmentation Architectures and Topological Constraints}
Extracting high-fidelity candidate regions demands architectures that balance local precision with global context. Early CNN-based models, such as U-Net \cite{ronneberger2015unet}, established an encoder-decoder paradigm that preserves spatial details via skip connections. However, their fixed, local receptive fields limit the ability to capture long-range continuity in elongated paths. The advent of Vision Transformers \cite{Dosovitskiy2020AnII} introduced global self-attention to model long-distance dependencies. To improve efficiency, hierarchical variants like Swin Transformer \cite{Liu2021SwinTH} and SegFormer \cite{Xie2021SegFormerSA} restricted attention to local windows. Nevertheless, these static window-based mechanisms remain suboptimal for the irregular, sparse geometry of path planning corridors. In contrast, the DAT \cite{xia2022vision} employs adaptive sampling to dynamically focus on informative regions. We are the first to employ DAT for candidate proposals, exploiting its sparse attention to track winding trajectories more effectively than fixed-window Transformers, while employing a deconvolutional decoder \cite{7410535} inspired by the U-Net paradigm to ensure pixel-level boundary precision. Consequently, CP-RPN is able to extract a narrower candidate region.

Furthermore, precise pixel classification does not guarantee structural connectivity. Standard objectives like cross-entropy \cite{rosenblatt1958perceptron} and dice loss \cite{Milletar2016VNetFC} optimize pixel-wise accuracy independently, often yielding fragmented predictions where slight discontinuities render a path infeasible. To address structural integrity, topological data analysis, specifically Persistent Homology \cite{Edelsbrunner2009ComputationalTA}, has been integrated into deep learning. Approaches like \cite{Hu2019TopologyPreservingDI} and \cite{Qi2023DynamicSC} introduced differentiable topological losses to penalize broken components in medical imaging, while graph-based constraints such as min-cut \cite{Xu2017DeepGF} and adjacency constraints \cite{Mosinska2017BeyondTP} explicitly model pixel connectivity. However, these methods focus on refining the overall topology while assigning equal importance to every fragment. This unguided strategy expends resources on unproductive dead ends. More importantly, it fails to guarantee a navigable start-to-goal path. Unlike prior works that focus on static shape preservation, our formulation enforces the connectivity of a valid start-to-goal trajectory, integrating topological persistence directly into the planning pipeline.

\section{METHOD}
\label{sec:method}

We propose the CP-RPN. This network explicitly maintains global topological consistency and preserves local boundary precision in complex environments. Consequently, our framework generates narrow and highly connected candidate regions. The proposed framework operates as a cohesive segmentation-guided planning pipeline as illustrated in Fig. \ref{fig_pipelne}. First, a DAT encoder processes the raw binary grid map. Its deformable attention mechanism dynamically captures long-range topological dependencies across the free space. These features are then progressively fused by a deconvolutional decoder to reconstruct a high-fidelity, topologically connected probability map. Finally, this continuous prediction is thresholded to form a binary candidate region, from which a Voronoi diagram is constructed to guide the deterministic path generation.

\subsection{Problem Formulation}
\label{sec:problem}

This work addresses the problem of 2D path planning for a point robot in a complex environment. The objective is to generate a feasible trajectory from a given start state \( x_s \) to a goal state \( x_g \) within a known map. The workspace is defined as \( \mathcal{X} \subset \mathbb{R}^2 \). Within this space, the obstacle space is denoted as \( \mathcal{X}_o \subset \mathcal{X} \), and the free space as \( \mathcal{X}_f = \mathcal{X} \setminus \mathcal{X}_o \). The goal region \( \mathcal{X}_{\text{goal}} \subset \mathcal{X}_f \) is a neighbourhood around the goal point \( x_g \) with a tolerance \( \epsilon \), formally defined as:

\begin{equation}
\mathcal{X}_{\text{goal}} = \{x \in \mathcal{X}_f \mid \|x - x_g\| \leq \epsilon \}.
\end{equation}

Instead of directly generating a path, CP-RPN first learns a segmentation model to predict a candidate free space region, denoted as \( \mathcal{X}_{\text{pred}} \), which is more likely to contain the optimal path. This strategy simplifies the path planning task by restricting the candidate region to a more promising area.

\subsection{Model Architecture}
\label{sec:model}
The architecture of CP-RPN is designed to identify thin, planning corridors within cluttered environments. We adopt a hybrid encoder-decoder structure, integrating a DAT \cite{xia2022vision} backbone for global context modeling with a deconvolutional \cite{ronneberger2015unet} decoding head for pixel-level reconstruction.

\textit{1) DAT encoder:}
Unlike standard convolutional neural networks that operate on fixed receptive fields, our encoder utilizes a hierarchical Transformer design to capture long-range dependencies. Given the spatial input configurations, the patch embedding layer tokenizes the image into a regular grid of disjoint patches. Following the standard hierarchical design \cite{xia2022vision}, the encoder processes features through four sequential stages. By progressively merging patches, the feature maps are downsampled by factors of $\{4, 8, 16, 32\}$ relative to the input resolution, while the channel dimensions are expanded to capture richer semantic information.

To efficiently model the environment, we employ a hybrid attention strategy as defined in our configuration. We alternate between Neighborhood Attention (NA), which focuses on local features within a fixed window to preserve obstacle boundaries, and Deformable Attention (DA), which captures global topology.

\textit{2) Deformable Self-Attention and Offset Learning:}
The deformable attention mechanism empowers the network to perceive winding paths by dynamically focusing on informative regions. As depicted in the attention module of Fig.\ref{fig_pipelne}, unlike standard self-attention that attends to a fixed grid, DA learns to adaptively shift its sampling locations toward informative regions.

Formally, let $q$ be a query element and $p_q$ be its reference point on the feature map. The network predicts a set of learnable offsets $\Delta p$ based on the query features. The attention output is computed by sampling features at these shifted locations:
\begin{equation}
\text{Attn}(q, x) = \sum_{k=1}^{K} A_{qk} \cdot x(p_q + \Delta p_{qk})
\end{equation}
here, $K$ denotes the number of sampling points, and $A_{qk}$ represents the normalized attention weight. The term $\Delta p_{qk}$ is the crucial offset generated by the offset network, which dynamically adjusts the sampling position $p_q + \Delta p_{qk}$ to align with the geometric structure of the candidate region. This allows the encoder to capture explicit connectivity information even across distant parts of the map.

\textit{3) Deconvolutional decoder and feature fusion:}
To recover the spatial resolution required for collision-free planning, we employ a deconvolutional decoder. Since deep encoder features contain strong semantic information but lack spatial precision, we employ deconvolution layers to actively reconstruct sharp boundaries while retaining the spatial recovery advantages of the U-Net architecture.

At each upsampling stage $i$, the spatial resolution of the feature map $F_{i-1}^{dec}$ is doubled via a transposed convolution operation:
\begin{equation}
F_{i}^{dec} = \text{ConvTranspose2d}(F_{i-1}^{dec})
\end{equation}
This fusion mechanism ensures that the final predicted candidate region $\mathcal{X}_{pred}$ maintains both the structural integrity captured by the learnable offsets and the precise boundaries required for safe path planning.

\subsection{Loss Function Design}
\label{sec:loss}

While standard pixel-wise losses like cross-entropy \cite{goodfellow2016deep} are effective for many segmentation tasks, they often fail to capture the critical structural properties required for path planning. Specifically, our task involves segmenting thin, sparse path corridors against a large background, which creates a severe class imbalance problem. This often leads to fragmented or discontinuous predictions that are unusable for a downstream planner.

To overcome these challenges, we introduce a composite loss function that combines pixel-level accuracy with explicit penalties for topological and connectivity errors. The total loss \( \mathcal{L}_{\text{total}} \) is defined as:

\begin{equation}
\mathcal{L}_{\text{total}} = \mathcal{L}_{\text{CE}} + \lambda_{\text{conn}} \mathcal{L}_{\text{conn}} + \lambda_{\text{topo}} \mathcal{L}_{\text{topo}},
\end{equation}

\noindent where \( \lambda_{\text{conn}} \) and \( \lambda_{\text{topo}} \) are weighting coefficients that balance the contributions of each loss term.

\subsubsection{Pixel-wise Cross-Entropy Loss ($\mathcal{L}_{\text{CE}}$)}

This is the foundational loss component, providing pixel-level supervision for the binary classification task. It is calculated as:

\begin{equation}
\mathcal{L}_{\text{CE}} = - \frac{1}{HW} \sum_{i,j=1}^{H,W} \sum_{c=1}^{C} y_{i,j}^{(c)} \log \hat{p}_{i,j}^{(c)},
\end{equation}

\noindent where \( H \) and \( W \) represent the height and width of the predicted output map, and \( C \) is the number of classes. The variable \( \hat{p}_{i,j}^{(c)} \) is the predicted probability for class \( c \) at pixel location \( (i,j) \), while \( y_{i,j}^{(c)} \) is the ground truth label, a one-hot vector. The factor \( \frac{1}{HW} \) normalizes the loss over all pixels.

\subsubsection{Connectivity-aware Loss ($\mathcal{L}_{\text{conn}}$)}

Local spatial coherence is indispensable for maintaining the structural continuity of predicted paths and preventing fragmentation \cite{Mosinska2017BeyondTP}. To enforce this prior, we integrate a connectivity-aware loss that penalizes local inconsistencies between neighboring pixels. This mechanism is formulated as a general local correlation objective:

\begin{equation}
\mathcal{L}_{conn} = -\frac{1}{HW} \sum_{i,j} \hat{p}_{i,j} \cdot \sigma \left( \frac{\text{corr}(\hat{p}_{i,j}, N(\hat{p}_{i,j}))}{\tau} \right),
\end{equation}

\noindent where $\hat{p}_{i,j}$ denotes the predicted probability of the free space class at pixel $(i,j)$, and $N(\hat{p}_{i,j})$ represents the mean probability within a local $3\times3$ neighborhood. The function $\text{corr}(\cdot)$ is employed to evaluate the local correlation between a pixel and its adjacent region, aiming to facilitate spatial consistency. By utilizing the sigmoid function $\sigma(\cdot)$ as an adaptive weighting mechanism with a temperature scale $\tau$, this loss term penalizes isolated point predictions and mitigates region fragmentation. The weighting factor $\lambda_{\text{conn}}$ is empirically set to 1.0 to balance the pixel-wise supervision.

\subsubsection{Topological Continuity Loss ($\mathcal{L}_{\text{topo}}$)}

This component incorporates persistent homology to maintain the global structural integrity of the predicted path. Persistent homology tracks the birth and death of topological features, such as connected components, as we build a filtration of the predicted probability map. We compute the Hausdorff distance between the persistence diagrams of the predicted segmentation and the ground truth. The objective is to minimize this distance, thereby enforcing topological consistency between the predicted path and the ground truth. The formulation is:

\begin{equation}
\mathcal{L}_{\text{topo}} = \frac{1}{B} \sum_{b=1}^{B} \mathcal{H}\left( \mathcal{D}(\hat{p}_b), \mathcal{D}(y_b) \right),
\end{equation}

\noindent where $B$ denotes the mini-batch size. For each sample $b$ in the batch, $\hat{p}_b$ and $y_b$ represent the predicted probability map and the ground truth, respectively. The function $\mathcal{D}(\cdot)$ extracts the persistence diagram, which is a multiset of points $(b, d)$ representing the birth and death times of topological features. In the $0$-th dimension, birth corresponds to the value at which a new connected component appears, while death denotes the value at which it merges with another. The Hausdorff distance $\mathcal{H}(\cdot, \cdot)$ between two diagrams effectively measures the discrepancy between the predicted topological structure and the ground truth. The weighting factor $\lambda_{\text{topo}}$ is set to 0.05. To balance computational efficiency, this loss is calculated every 10 batches during training.

\subsection{Path Generation}
Once the network predicts the candidate region $\mathcal{X}_{pred}$, the final trajectory is generated using a deterministic geometric approach. Specifically, we extract the topological skeleton of the region via a Voronoi diagram \cite{Choset1996SensorBM}. This formulation transforms the continuous free space into a sparse graph that intrinsically maximizes obstacle clearance. Following a global search on this skeleton, we employ a localized A* refinement to guarantee discrete connectivity and strictly collision-free transitions within narrow bottlenecks, thereby compensating for discretization artifacts via dilation without altering the global topology.

\section{Experiments and Results}

\subsection{Experimental Setup}

\subsubsection{Datasets and Metrics}
Experiments are conducted on a Random Forest binary map dataset. The dataset contains 2,500 unique environments, each paired with 25 different starting positions and corresponding ground-truth paths. The ground-truth solutions are generated using the RRT* algorithm \cite{Karaman2011SamplingbasedAF} through extensive iterations to ensure near-optimality. Our experiments are exclusively run on maps with a resolution of $480 \times 480$. Unlike conventional grid maps\cite{Wang2020NeuralRL}\cite{Qureshi2018MotionPN}, our dataset simulates dense and sparse forest-like environments. All our models are trained exclusively on a subset of the dense maps. The models are then evaluated on a testing set consisting of unknown maps, including the sparse maps, to assess the models’ generalization capabilities.

\subsubsection{Implementation Details}
Our model adopts a segmentation-based framework integrated with the DAT \cite{xia2022vision} backbone. Leveraging these candidate regions, we generate the final trajectory using a Voronoi-based planner with an A fallback; despite this hybrid process, the system maintains exceptional speed due to the highly constrained search space. The network is trained for 14 epochs with a batch size of 4. Optimization is performed using the Adam optimizer \cite{Kingma2014AdamAM} with $\beta_1=0.9$, $\beta_2=0.98$, and a dynamic learning rate schedule with 3200 warmup steps, facilitating rapid convergence. Weight decay is set to $1 \times 10^{-5}$ to prevent overfitting \cite{Loshchilov2017DecoupledWD}. All experiments are conducted on $480 \times 480$ resolution maps. Training is conducted on two NVIDIA RTX 3090 GPUs (24 GB each). Our training process takes roughly 15.4 hours. 

\subsection{Comparative Experiments}

We evaluate CP-RPN in dense simulated environments against a set of learning-based and traditional motion planning baselines. The primary baseline is MPT \cite{johnson2021motion}, which predicts navigable regions for downstream planners. We also evaluate MPNet \cite{Qureshi2018MotionPN}, a learning-based motion planning approach using neural networks for path generation. For classical methods, we adopt RRT \cite{LaValle1998RapidlyexploringRT} and Informed RRT* \cite{Karaman2011SamplingbasedAF}. These baselines are tested under the same map configurations and $480 \times 480$ resolution.

\begin{table}[t!]
    \centering
    \caption{Comparison of different path planning methods}
    \label{tab:comparison}
    \begin{tabular}{lccc}
        \hline
        \textbf{Method} & \textbf{Success Rate (\%)} & \textbf{Time (s)} & \textbf{Region Size (Pixels)} \\
        \hline
        RRT* & 100.00 & 5.47 & - \\
        IRRT* & 100.00 & 0.34 & - \\
        UNet-RRT* & 30.27 & 0.18 & - \\
        MPNet & 92.35 & 0.29 & - \\
        MPT-RRT* & 99.40 & 0.19 & 18029 \\
        \textbf{Ours-voronoi} & \textbf{99.60} & \textbf{0.11} & \textbf{7187} \\
        \hline
    \end{tabular}
    \vspace{-0.4cm}
\end{table}

\begin{figure}[t!]
    \centering
    \includegraphics[width=1\linewidth]{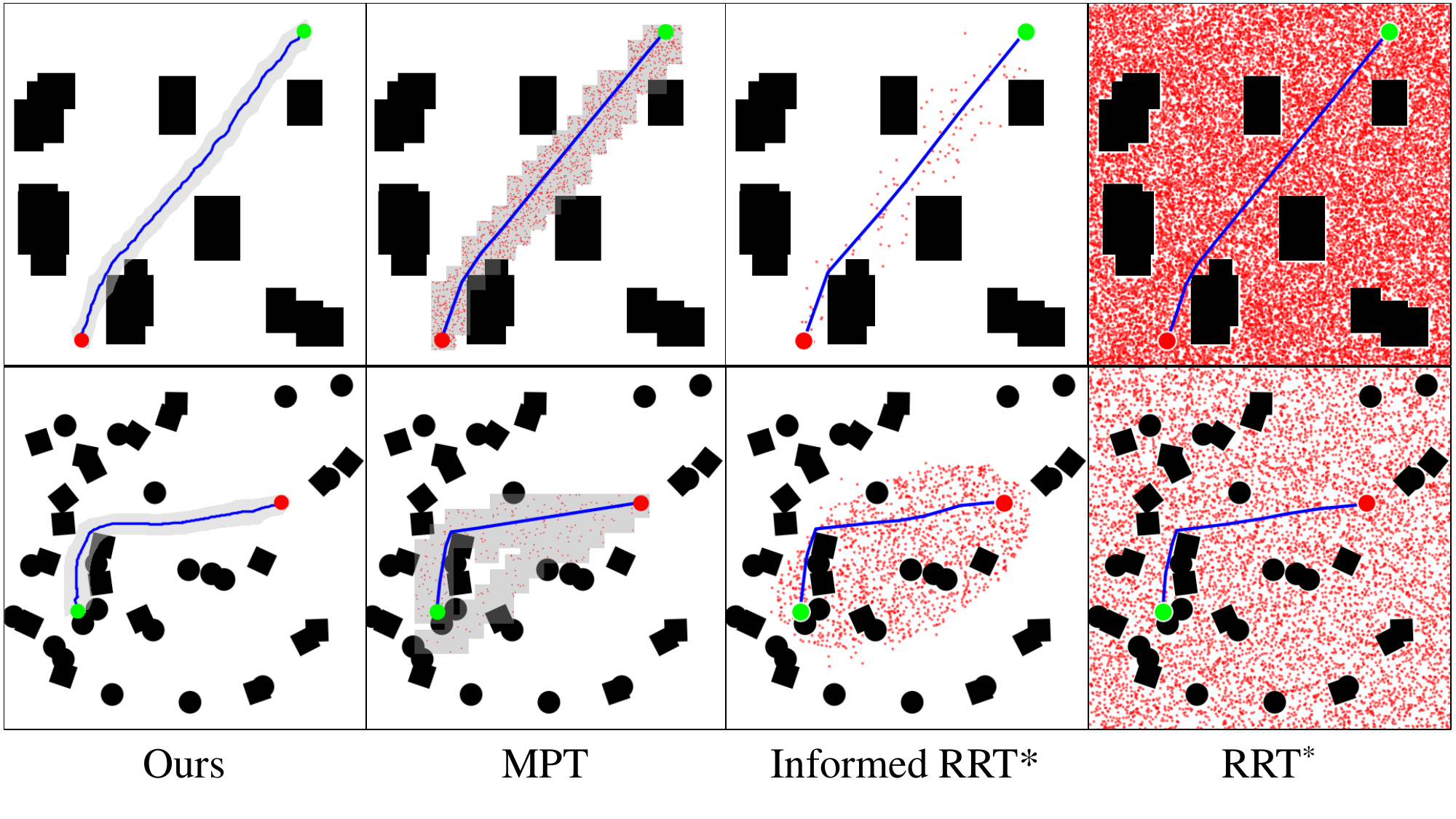}
    \vspace{-0.8cm}
    \caption{Qualitative comparison of path planning results. From left to right: Ours, MPT, IRRT*, and RRT*. The red dot represents the start, and the green dot represents the goal.}
    \label{fig:qualitative_results}
\end{figure}

As summarized in Table \ref{tab:comparison}, CP-RPN achieves a success rate of 99.60\%, which is highly competitive with the MPT-RRT* baseline (99.40\%) while significantly outperforming other learning-based methods like MPNet (92.35\%). Qualitatively, as shown in Fig. \ref{fig:qualitative_results}, traditional sampling-based methods like RRT* often explore unnecessary regions before convergence. The baseline MPT, while restricting the search space, tends to predict overly broad or disconnected regions (as seen in the second column), which can trap the downstream optimizer. In contrast, CP-RPN predicts a highly compact and topologically connected path region that closely wraps around the optimal trajectory, significantly reducing the collision-checking overhead for the subsequent Voronoi-based planner.

\begin{figure}[t!]
    \centering
    \includegraphics[trim={0 1.9cm 0 0}, clip,width=1.0\linewidth]{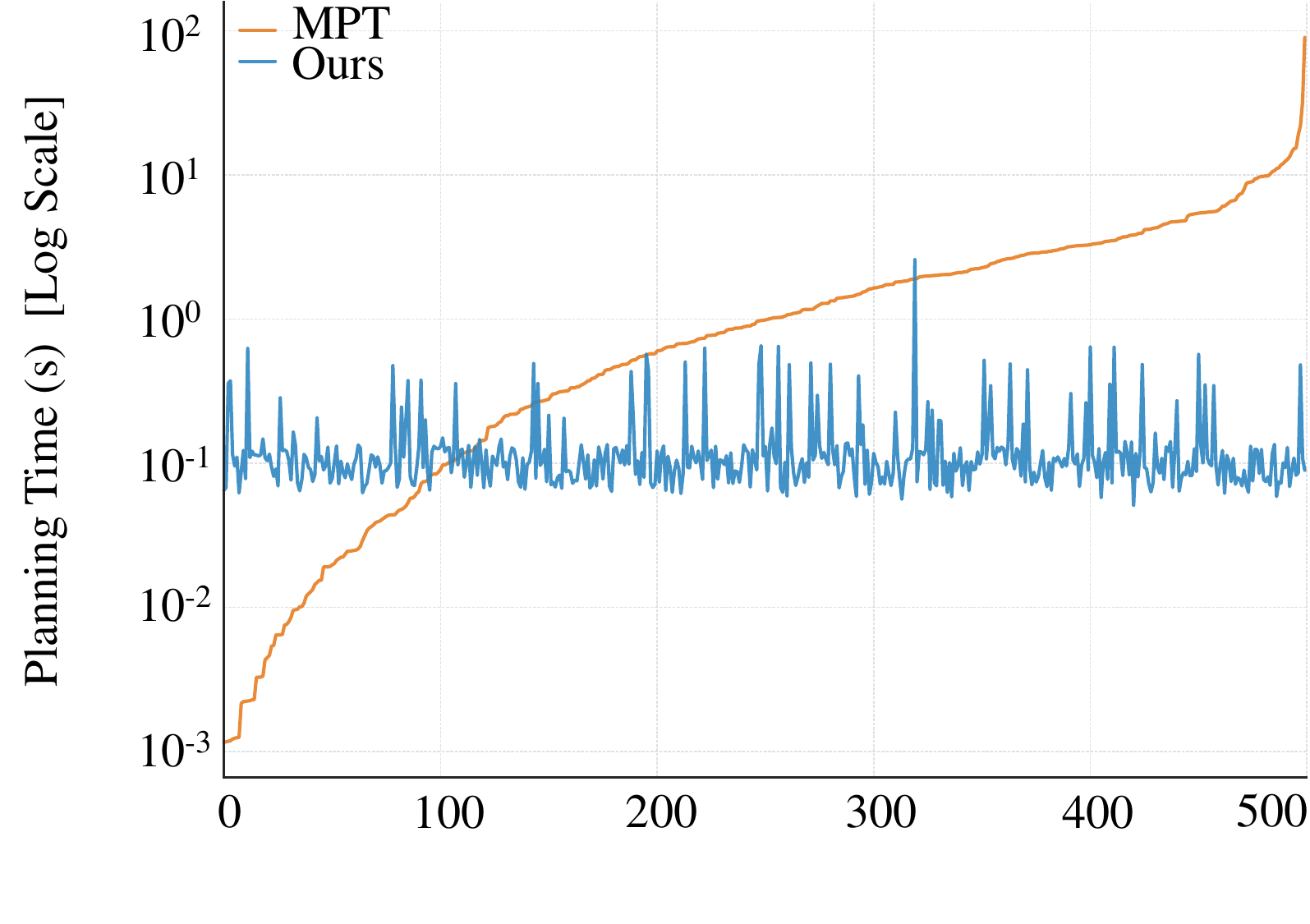}
    \vspace{-0.4cm}
    \caption{Planning latency comparison. The environments are sorted by the computational cost of the MPT baseline (orange) to represent increasing difficulty. The blue curve denotes the planning time of our method. The y-axis is plotted on a logarithmic scale.}
    \label{fig:complexity}
    \vspace{-0.4cm}
\end{figure}

To rigorously evaluate the scalability of CP-RPN, we sorted 500 validation environments based on the computational difficulty experienced by the MPT baseline. Fig. \ref{fig:complexity} reveals a critical limitation in the baseline method: MPT (orange curve) exhibits a steep monotonic increase in latency as the environment becomes more complex, indicating a strong positive correlation between computational cost and obstacle density. In sharp contrast, CP-RPN (blue curve) demonstrates a remarkable characteristic of complexity decoupling. The planning time remains nearly constant ($\approx 0.1s$) regardless of the topological difficulty. This time-invariance property is crucial for safety-critical robotic systems, as it prevents unexpected latency spikes in complex unseen environments.

\begin{figure}[t!]
    \centering
    \hspace{-0.5cm} 
    \includegraphics[width=1.0\linewidth]{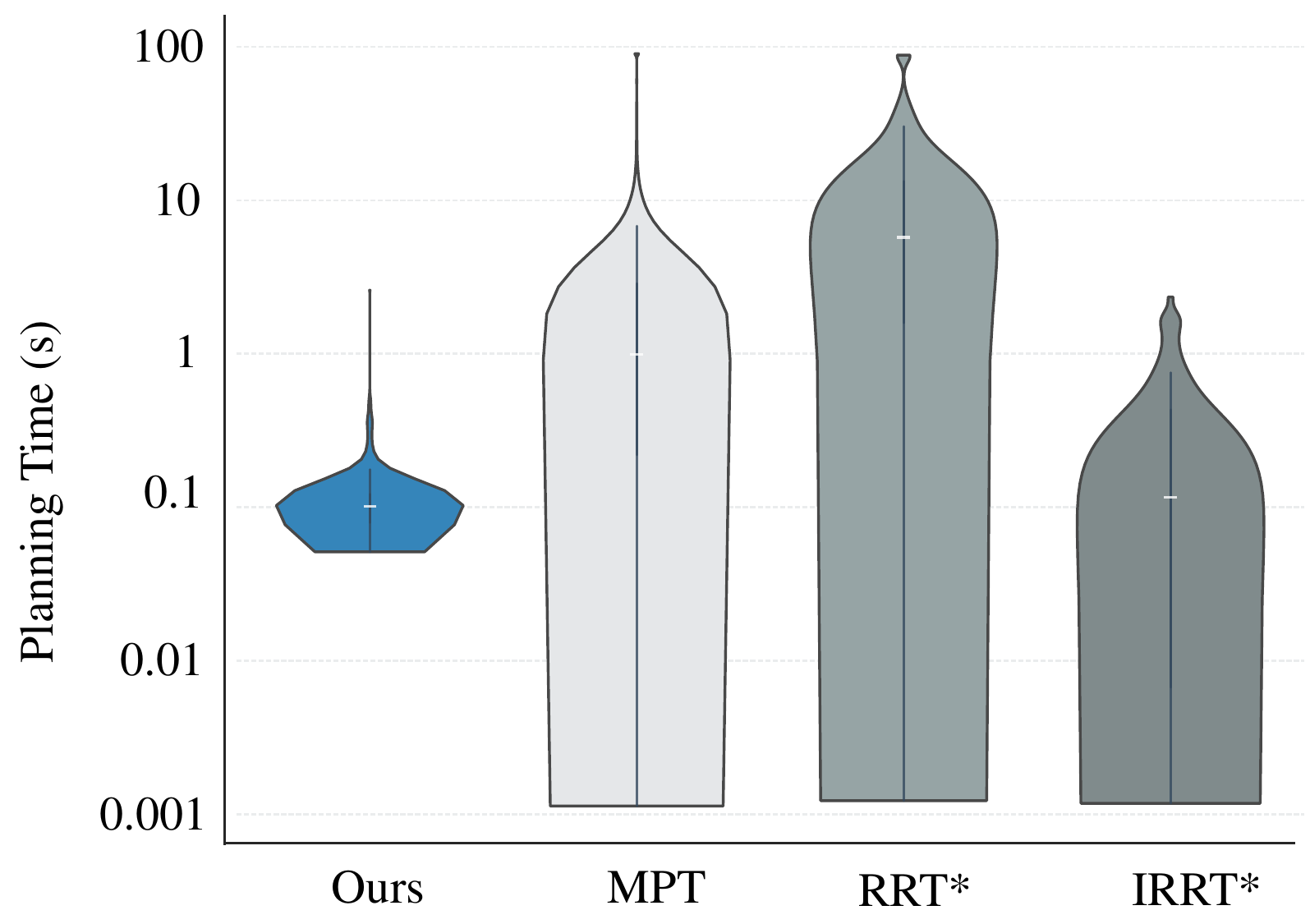}
    \vspace{-0.2cm} 
    \caption{Time stability and robustness analysis. The heavy-tailed distribution of RRT* and Informed-RRT* indicates high variance and potential timeout risks. In contrast, the compact distribution of our method indicates \textbf{deterministic performance}, suitable for real-time control loops.}
    \label{fig:stability}
    
\end{figure}

Beyond average efficiency, the stability of the planning algorithm is paramount for real-time control loops. Fig. \ref{fig:stability} compares the planning time distributions using a logarithmic-scale violin plot. Traditional sampling-based methods (RRT*, Informed-RRT*) exhibit a distinct heavy-tailed distribution, where elongated upper tails indicate high variance and unpredictable worst-case execution times. Conversely, CP-RPN manifests a highly compact distribution with negligible variance. The absence of outliers confirms that CP-RPN provides deterministic latency guarantees, eliminating the probabilistic uncertainty inherent in random sampling and ensuring the robot can consistently replan at a fixed frequency.

\begin{figure}[t!]
    \centering
    \includegraphics[width=1.0\linewidth]{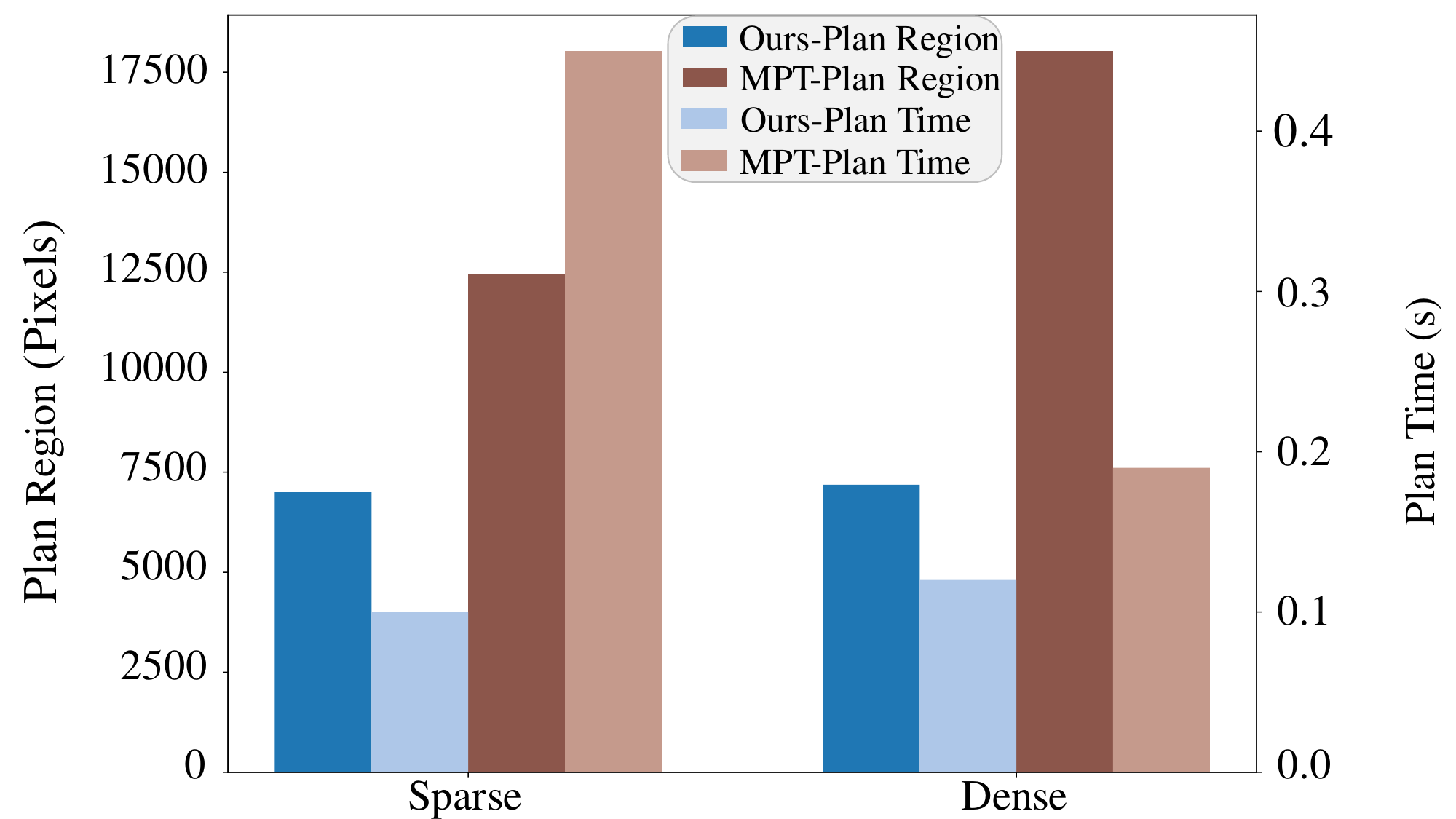}
    \vspace{-0.6cm}
    \caption{Comparison of planning time and area for CP-RPN and MPT-RRT* in sparse and dense environments. CP-RPN achieves significantly smaller candidate regions and faster planning times in both settings.}
    \label{fig:time_area}
    \vspace{-0.4cm}
\end{figure}

In addition to temporal efficiency, CP-RPN exhibits superior spatial compactness. As shown in Fig. \ref{fig:time_area} and Table \ref{tab:comparison}, CP-RPN maintains a consistently low planning time in both sparse (0.10s) and dense (0.11s) environments, whereas MPT-RRT* slows down significantly in sparse settings (0.45s). More importantly, our model demonstrates a remarkable advantage in planned area efficiency. CP-RPN predicts a significantly smaller planned region compared to MPT-RRT*. This indicates that CP-RPN generates a highly refined navigable space that is sufficient for planning but free from excessive redundant areas.

\begin{figure}[t!]
    \centering
    \includegraphics[width=1\linewidth]{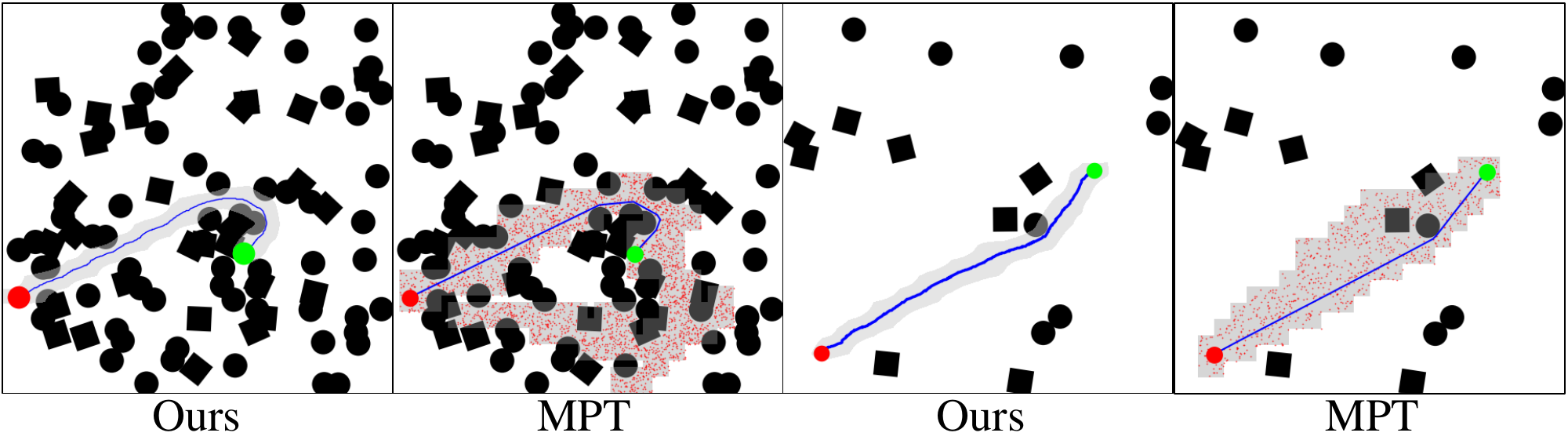}
    \vspace{-0.7cm}
    \caption{Generalization capability across different obstacle densities. The left row shows results in dense environments, while the right row shows sparse environments.}
    \label{fig:sparse_dense}
    \vspace{-0.4cm}
\end{figure}

Finally, to validate the generalization capability of our model, we evaluated it on environments with varying obstacle densities that were not seen during training. Fig. \ref{fig:sparse_dense} illustrates the segmentation results in dense forests versus sparse forests. Although trained primarily on dense maps, the network successfully adapts to sparse environments. It does not blindly predict large free spaces; instead, it utilizes the DAT backbone's global context awareness to identify the most efficient path region connecting the start and goal. This adaptability ensures that CP-RPN remains robust and efficient regardless of the obstacle density.

\subsection{Ablation Study}

To assess each module's contribution, we conducted ablation experiments in dense environments. Connectivity accuracy is the primary metric, which is particularly stringent given our significantly narrower ground-truth masks compared to existing baselines.

\begin{table}[h]
    \centering
    \caption{ABLATION STUDY OF THE PROPOSED CP-RPN}
    \label{tab:ablation_study}
    \begin{tabular}{lcc}
        \hline
        \textbf{Method} & \textbf{Accuracy (\%)} & \textbf{Time (s)} \\
        \hline
        w/o DAT \& $\mathcal{L}_{conn}$ \& $\mathcal{L}_{topo}$ & 87.31 & 0.13 \\
        w/o $\mathcal{L}_{conn}$ \& $\mathcal{L}_{topo}$ & 95.83 & 0.13 \\
        w/o deconvolutions decoder & 98.68 & 0.14 \\
        \textbf{Ours (Full Model)} & \textbf{99.60} & \textbf{0.11} \\
        \hline
    \end{tabular}
\end{table}

As shown in Table \ref{tab:ablation_study}, our full architecture achieves 99.60\% accuracy with 0.11s latency. Replacing the DAT backbone with a standard Transformer (dropping accuracy to 87.31\%) suggests that static attention struggles with the irregular, thin structures of narrow corridors. Conversely, DAT's deformable receptive fields adaptively sample informative features along obstacle boundaries, facilitating robust global topology capture.

The proposed auxiliary losses are critical for maintaining structural integrity. Removing $\mathcal{L}_{conn}$ and $\mathcal{L}_{topo}$ reduces accuracy to 95.83\% and results in noticeable region fragmentation. While the DAT backbone captures global context, these auxiliary terms provide necessary structural constraints: $\mathcal{L}_{conn}$ enforces local consistency through neighbor-wise correlation to facilitate smooth and coherent path segments, while $\mathcal{L}_{topo}$ preserves global connectivity via persistent homology. This integration ensures that the predicted regions remain structurally continuous from the start to the goal.

The cascaded deconvolutional decoder is essential for spatial precision. Removing this module reduces accuracy to 98.68\% and increases planning time to 0.14s. This indicates that while the Transformer layers model long-range dependencies, an encoder-decoder architecture following the U-Net paradigm is critical to recover fine-grained spatial details. Consequently, these higher-fidelity masks allow the model to effectively prune redundant free space, accelerating the downstream Voronoi-based planning by narrowing the search area.

\subsection{Real-World Experiments}
\begin{figure}[t!]
    \centering
    \includegraphics[width=1\linewidth]{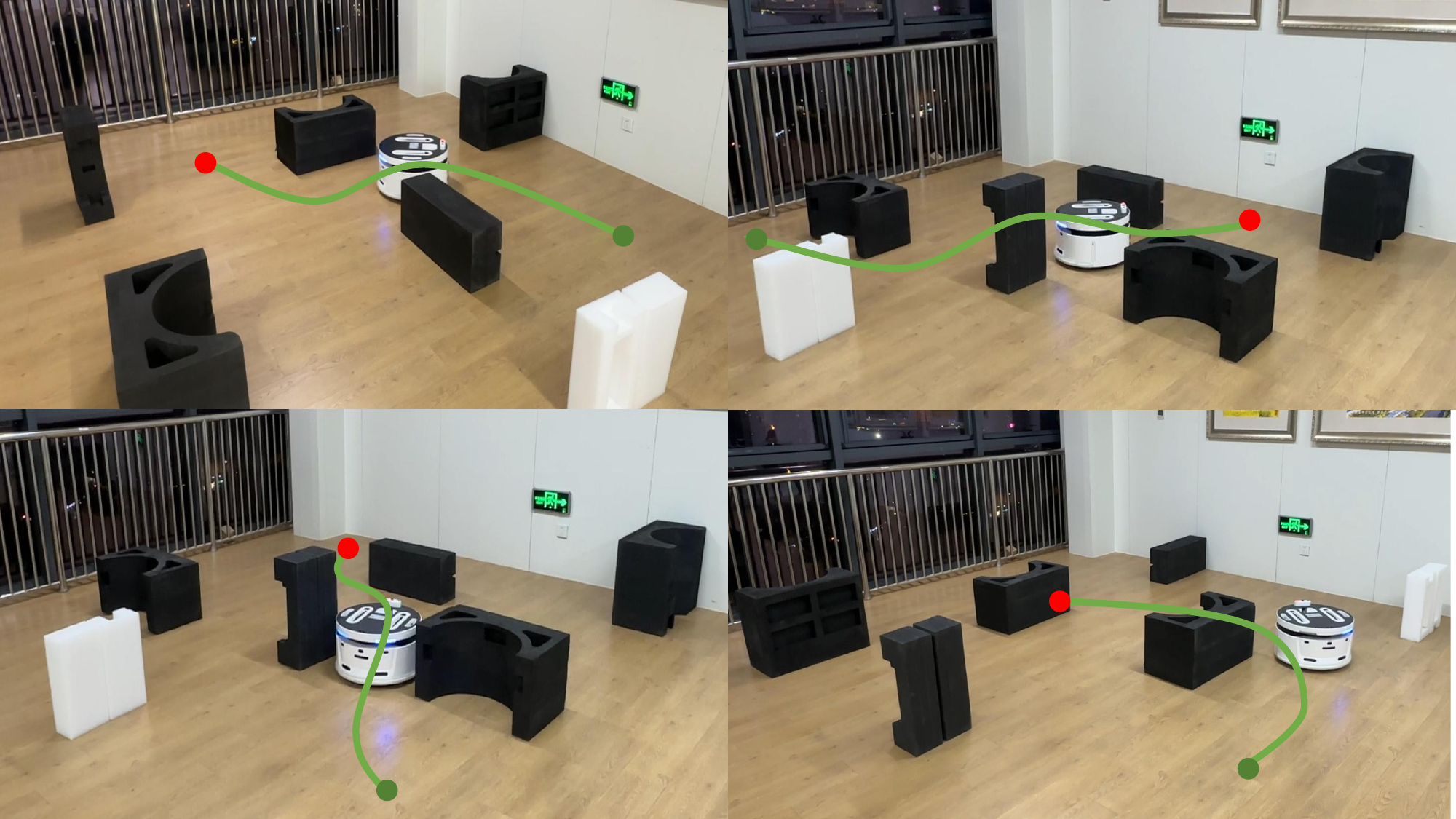}
    \vspace{-0.6cm}
    \caption{Real-world experiments of CP-RPN in a cluttered indoor environment. The remote IPC performs inference and global planning, and the robot reaches all targets without collisions.}
    \label{fig:real-world}
    \vspace{-0.4cm}
\end{figure}
To validate the sim-to-real transferability of our proposed CP-RPN, we conducted physical experiments in a cluttered indoor environment populated with irregular obstacles, as shown in Fig.\ref{fig:real-world}. We adopted a host-client architecture for deployment: the deep learning model and global planner were executed on a remote Industrial PC to ensure real-time inference, while the mobile chassis handled low-level execution. Our evaluation was conducted across three different environments, testing three distinct start-goal configurations in each. To ensure robust navigation and safety in the physical world, we made specific adjustments for real-world deployment: we activated the chassis's onboard local obstacle avoidance algorithm and increased the inflation radius in the planning parameters. The robot successfully navigated to the designated targets in all tests without collisions. These results demonstrate that CP-RPN can effectively guide mobile robots in real-world scenarios, bridging the gap between simulated training and physical deployment.

\section{Conclusion}
In this work, we presented CP-RPN, a connectivity-preserving region proposal network designed to accelerate robot path planning. By integrating a DAT backbone for global context modelling with a deconvolutional decoder following the U-Net paradigm, our framework effectively maintains spatial precision while preserving the topological integrity of candidate regions. Experimental results demonstrate that CP-RPN achieves 99.60\% pixel-wise accuracy and reduces the search space by 60.13\%, enabling downstream planners to find valid paths within 0.11s consistently. While the Voronoi-based skeletons accelerate path planning, they inherently exhibit geometric jaggedness. Future work will thus focus on embedding curvature-based smoothing constraints directly into the topological loss to further improve path smoothness.

\bibliographystyle{IEEEtran}
\bibliography{ref}

\end{document}